\documentclass{Interspeech}

\usepackage{tipa}
\usepackage{graphicx}
\usepackage{subcaption}
\usepackage{caption}
\usepackage{soul}
\usepackage{xcolor}
\usepackage{multirow}

\usepackage{array}
\newcolumntype{?}{!{\vrule width 1pt}}

\usepackage{numprint}
\npthousandsep{\,}

\usepackage{amsmath}
\usepackage{todonotes}
\usepackage{booktabs}

\usepackage[commandnameprefix=ifneeded, todonotes={textsize=small}]{changes}
\definechangesauthor[color=teal]{df}
\definechangesauthor[color=blue]{mn}
\definechangesauthor[color=olive]{mg}
\definechangesauthor[color=red]{mc}

\newcommand{\df}{\textcolor{black}}

\newcommand{\phoneme}{\Phi}
\newcommand{\phone}{\phi}

\interspeechcameraready

\title{Feature Attribution Unveils Parallels \\ between Relevant Acoustic Cues in ASR and X}

\title{Echoes of X: \\ Unveiling Relevant Acoustic Cues in ASR through Feature Attribution}

\title{Unveiling Relevant Acoustic Cues in ASR through Feature Attribution}

\title{Echoes of Phonetics: \\ Unveiling Relevant Acoustic Cues for ASR via Feature Attribution}

\author[affiliation={1,2}]{Dennis}{Fucci}
\author[affiliation={2}]{Marco}{Gaido}
\author[affiliation={2}]{Matteo}{Negri}
\author[affiliation={2}]{Mauro}{Cettolo}
\author[affiliation={2}]{Luisa}{Bentivogli}

\affiliation{Department of Information Engineering and Computer Science}{University of Trento}{Italy}
\affiliation{Machine Translation Unit}{Fondazione Bruno Kessler}{Italy}
\email{dfucci@fbk.eu}

\begin{document}

\maketitle

\begin{abstract}
Despite significant advances in ASR, the specific acoustic cues models rely on remain unclear.
Prior studies have 
examined
such cues on a limited set of phonemes and outdated models. In this work, we apply a feature attribution technique to identify the relevant acoustic cues for a modern Conformer-based ASR system. 
By analyzing plosives, fricatives, and vowels, we assess how feature attributions align with their acoustic properties in the time and frequency domains, also 
essential
for human speech perception.
Our findings show that the ASR model relies on vowels’ full time spans, particularly their first two formants, with 
greater 
saliency in male speech. 
It also better captures the spectral characteristics of sibilant fricatives than non-sibilants and prioritizes the release phase in plosives, especially burst characteristics. 
%
%
%
%
These insights enhance the interpretability of ASR models and highlight areas for future research to 
uncover
potential gaps in model robustness.
\end{abstract}

\section{Introduction}

State-of-the-art neural automatic speech recognition (ASR) systems achieve impressive performance across languages but remain opaque, offering limited insight into how they process acoustic information. 
To fill this gap, researchers have
recently explored
how speech models organize phonemic information, investigating their hidden 
states
through 
probing classification 
(e.g., \cite{belinkov19_interspeech, deseyssel22_interspeech, cormac-english-etal-2022-domain, martin23_interspeech, choi24b_interspeech}),
discrimination (e.g., \cite{alishahi-etal-2017-encoding, chaabouni17_interspeech}), and 
clustering (e.g., \cite{nagamine_2017, krug2018}).
Fewer studies \cite{ravanelli_sincnet, krug-stober-2018-introspection, palaz_filters, krug2020, trinh21_bubble, fucci-spes}, instead, have explored how 
ASR models use acoustically distinctive cues that are also perceptually relevant to humans.

These acoustic properties vary according to the phoneme category.
For instance, vowel quality is primarily determined by steady-state formant frequencies, especially the first two, which indicate tongue height and backness (e.g., \cite{hillenbrand_vowels}).
Plosives are characterized by several cues, among which the release of energy after the closure is essential for distinguishing the place of articulation---i.e., the burst---
(e.g., \cite{revoile_burst, jackson_plosives}).
Fricatives are characterized by frication noise, with specific spectral properties varying by place of articulation 
(e.g., \cite{jongman_fricatives, maniwa_fricatives}).
%
%
Regarding
these characteristics, prior studies in ASR only offer partial insights.  Analyzing convolutional filters in hybrid HMM-neural and neural models, \cite{ravanelli_sincnet} and \cite{palaz_filters} observed frequency response peaks in regions corresponding to fundamental frequency and the first two formants of vowels. 
\cite{krug-stober-2018-introspection} and \cite{krug2020} found that 
a 
CTC-based
convolutional acoustic model 
detects signals in frequency ranges associated with the first two formants for predicting /\textipa{a}/ and /\textipa{\ae}/, and burst signals for /\textipa{t}/.
Using a feature attribution technique that assigns saliency scores to input features by perturbing the audio input, \cite{trinh21_bubble} demonstrated that an acoustic model combining time-delay neural networks and LSTM, unlike a GMM-based acoustic model, attends to high-energy regions in spectrograms.
Using a similar approach, \cite{fucci-spes} found that a Conformer-based model focuses on formant ranges for /\textipa{o\textupsilon}/ and distinctive noise ranges for /\textipa{s}/ and /\textipa{n}/.
%
%
%
%
%
Despite 
these contributions 
to model interpretability, 
prior works
have only
examined outdated architectures and/or conducted limited qualitative analyses on a narrow 
set of phonemes.

To address these gaps, this work provides 
the 
first fine-grained analysis on 
\textit{a 
\df{wider}
range of phonemes}, emphasizing the impact of their primary acoustic cues on 
predictions
by \textit{a modern autoregressive ASR model}.
We examine vowels, fricatives, and plosives in English---a language where ASR systems excel and which benefits from extensive acoustic-phonetic research.
Using a Conformer encoder-Transformer decoder model \cite{gulati20_interspeech}---a widespread architecture
in state-of-the-art models---and a feature attribution technique, 
we assess how saliency scores distribute across phoneme duration and 
the extent to which
they 
highlight
key cues like vowel formants, frication noise, and plosive release.
Our analysis shows that the model assigns varying emphasis to 
phonetic time spans---maximally for vowels and minimally for plosive closures.
In the frequency domain, it relies on cues characteristic of the analyzed phonemes and crucial for human perception, such as the first two formants in vowels and spectral peaks in fricatives and plosive releases. This is less pronounced for phonemes with less distinctive spectral cues, like non-sibilant fricatives and bilabial plosives.
Additionally, gender differences emerge, with stronger alignment, for example, between salient features and the first two formants in men's vowels.
%
%
%
%
Our findings enhance ASR model interpretability, revealing mechanisms deeply rooted in 
phonetic knowledge.\footnote{\df{We release the code under the Apache 2.0 at \url{https://github.com/hlt-mt/phonetic-analysis-xai}}.}

\section{Method}

\begin{figure}[t!]
    \centering
    \includegraphics[width=0.42\textwidth]{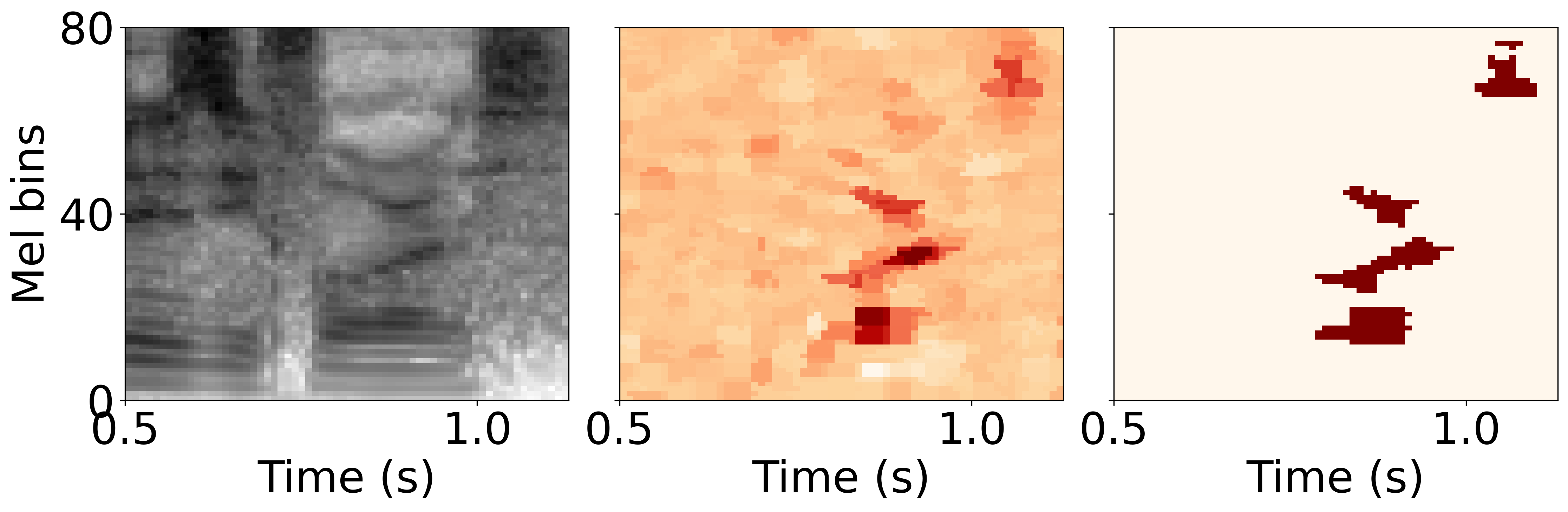}
    \caption{Mel-spectrogram (left), saliency map (center), and binary map (right) for the predicted token ``us''.}
    \label{fig:s_example}
\end{figure}

Given an ASR model that takes a mel-spectrogram \( \mathbf{X} \in \mathbb{R}^{T \times F} \) as input---where \( T \) is the number of time frames and \( F \) the number of frequency bins---and 
autoregressively
predicts a token sequence \( \mathbf{y} = (y_0, y_1, \ldots, y_I) \) 
of length \( I \)
based on \( \mathbf{X} \) and prior tokens \( (y_0, \ldots, y_{i-1}) \), we investigate the relationship between \( \mathbf{X} \) and each predicted token \( y_i \).
%
To this end, we use a feature attribution method to assign a saliency score to each element of \( \mathbf{X} \), 
providing a measure of its importance in generating \( y_i \).
This results in a \textit{saliency map} \( \mathbf{S}_i \in \mathbb{R}^{T \times F} \) which highlights the spectrogram regions most relevant for predicting
\( y_i \).
Specifically, we employ SPES \cite{fucci-spes}, 
a technique tailored to autoregressive speech-to-text 
models 
that has shown superior performance compared to previous ASR feature attribution methods.
SPES clusters input spectrogram 
elements
by energy, isolating features like harmonics and noise, and quantifies the impact of perturbing these clusters 
on \( y_i \).
%
Finally, we convert \( \mathbf{S}_i \) into a binary map \( \mathbf{S'}_i \in \{0, 1\}^{T \times F} \), retaining only
the most salient elements 
in \( \mathbf{S}_i \)
(see Fig. \ref{fig:s_example}). 
Based
on manual inspection of the saliency distribution, we retain the top $3$\% of elements, as they best capture salient 
features.

We analyze the explanations \( \mathbf{S'} \) of tokens containing phonemes of three common classes:
\textbf{vowels} (/\textipa{\textscripta}/, /\textipa{\ae}/, /\textipa{\textturnv}/, /\textipa{\textepsilon}/, /\textipa{\textschwa}/, /\textipa{\textupsilon}/, /\textipa{\textsci}/, /\textipa{\textbari}/, /\textipa{i}/, /\textipa{\textbaru}/, /\textipa{u}/),  
\textbf{plosives} preceding a vowel (/\textipa{p}/, /\textipa{b}/, /\textipa{k}/, /\textipa{g}/, /\textipa{t}/, /\textipa{d}/),  
and \textbf{fricatives} (/\textipa{s}/, /\textipa{z}/, /\textipa{\textesh}/, /\textipa{\textyogh}/, /\textipa{f}/, /\textipa{v}/, /\textipa{\texttheta}/, /\textipa{\dh}/).
Within \( \mathbf{S'}_i \), we focus on the time frames where each phoneme is produced
%
analyzing how the saliency scores are distributed in time and frequency, and verifying whether salient elements correspond to 
distinctive acoustic cues.

\noindent \textbf{Time.} 
To 
assess
the extent to which the model attends to an instance (\(\phone\)) of a phoneme (\(\phoneme\)),
we compute the \textit{time coverage} (\(\text{TC}_\phone\)), defined as the percentage of frames in \( \mathbf{S'}_i \) during \(\phone\)'s duration 
with at least one salient element along the frequency axis. For plosives, we analyze 
closure and release phases separately.

\noindent \textbf{Frequency.}
For the frequency analysis, we assess whether the salient elements in \( \mathbf{S'}_i \) correspond to frequency values of distinctive spectral properties of
$\phoneme$.
We extract steady-state spectral properties from the input speech: for vowels, we measure the first four formants (F1–F4) at the midpoint;\footnote{Formant measurements are carried out using Parselmouth \cite{parselmouth}. Formant ceilings are set to \numprint{5500} Hz for women and \numprint{5000} Hz for men, with five formants, a time step of $10$ ms, and a window size of $25$ ms.} 
for fricatives, we extract the spectral peak location at the midpoint;
for
plosives, we determine the spectral peak location at the onset of the release phase to capture burst characteristics.\footnote{
Spectral peak locations are extracted 
from the input mel-spectrograms by identifying the frequency bin with the highest value.
For
burst, the two initial frames---a $35$ ms window---are considered.}
Then, for each phoneme $\phoneme$, we compute the \textit{spectral match} (\( \text{SM}_\phoneme \))---the percentage 
of formant frequencies and spectral peaks that correspond to salient elements in \( \mathbf{S'}_i \)---quantifying 
how often these acoustic properties are salient for model's prediction.
For a broader perspective on saliency score distribution and to assess 
whether
high scores also appear in frequency regions beyond the extracted spectral cues, we analyze the \textit{distribution of 
salient elements
across all frequencies} for all occurrences of $\phoneme$ (\( \text{D}_\phoneme \)).

\section{Experimental Setting}
\label{sec:exp_sett}

\noindent \textbf{ASR Model.}
We train an autoregressive encoder-decoder ASR model that processes log-compressed mel-filterbank features ($80$ channels), computed over $25$ ms windows with a $10$ ms stride using PyKaldi \cite{pykaldi}.
Output text is
encoded into BPE \cite{sennrich-etal-2016-neural} with a vocabulary size of \numprint{8000}, using SentencePiece \cite{kudo-richardson-2018-sentencepiece}.
The input features are normalized via utterance-level Cepstral Mean and Variance Normalization and downsampled by a factor of $4$ through two $1$D convolutional layers with a stride of $2$. The processed features are then passed to a $12$-layer Conformer encoder \cite{gulati20_interspeech} and a $6$-layer Transformer decoder \cite{vaswani}. The 
encoder layers use 
a 
convolution kernel size of $31$,
an embedding size of $512$, and a linear layer hidden size of \numprint{2048}. 
The model
comprises $113$M parameters and is implemented using the fairseq-S2T framework \cite{wang-etal-2020-fairseq}.
Training is conducted on CommonVoice \cite{commonvoice}, LibriSpeech \cite{librispeech}, TEDLIUM v3 \cite{tedlium}, and VoxPopuli \cite{voxpopuli}, for \numprint{250000} steps by minimizing a label-smoothed cross-entropy loss \cite{szegedy-etal-2016} on the decoder output, augmented with an auxiliary CTC loss \cite{Graves2006ConnectionistTC} 
computed on 
the output of the $8$th encoder layer. Optimization is performed with 
Adam
\cite{kingma-ba-2015} ($\beta_1=0.9$, $\beta_2=0.98$) and a Noam learning rate scheduler \cite{vaswani} 
with
a learning rate peak
of $0.002$ 
and \numprint{25000} warmup steps.
During training, we apply SpecAugment \cite{specaugment} with a dropout probability of $0.1$. The final model is obtained by averaging the last 
seven
checkpoints.

\noindent \textbf{Data.}
We conduct our experiments on TIMIT \cite{timit}, a dataset featuring time-aligned transcriptions of read speech from both male and female speakers of American English.
TIMIT is chosen for 
its rich phonetic inventory and annotation,
and has been used in other studies aiming to interpret speech models (e.g., \cite{ravanelli_sincnet, belinkov19_interspeech, cormac-english-etal-2022-domain}).
We specifically focus on the SX subset of TIMIT, which contains $450$ sentences designed to provide 
robust variety of phonemes.
Each sentence is read by seven speakers, totaling \numprint{3150} utterances. 
\df{On this SX set, our ASR model achieves a WER of $6.69$ ($6.10$ when considering only the test portion),}\footnote{Calculated with JiWER 3.0.5 (\url{https://github.com/jitsi/jiwer}), with punctuation removed and text lowercased.}
and produces error-free transcriptions for \numprint{2191} sentences. 
To avoid bias from mismatched predictions and annotations, we focus on the error-free predictions.

\noindent \textbf{Feature Attribution.}
SPES's hyperparameters follow 
\cite{fucci-spes}. 
Masking is performed over \numprint{20000} iterations, with each segment randomly masked at a probability of $0.5$.
After generating saliency maps for all tokens in a predicted sentence, we apply mean-standard normalization to each explanation individually.
Since tokens in \( \mathbf{y} \) correspond to subwords, as in most state-of-the-art ASR models, but TIMIT provides segmentations for phonemes and words, 
we aggregate the saliency maps for a word 
consisting
of multiple tokens by taking the element-wise maximum across all tokens. 
This approach captures the most salient elements from all tokens forming the word, enabling us to align each predicted word with its corresponding segmentation. From this segmentation, we trace back to the phonetic level 
and isolate the \( \mathbf{S'}_i \) portions corresponding to the $\phone$ spans.


\begin{figure*}[t!]
    \centering
    \includegraphics[width=0.85\textwidth]{
    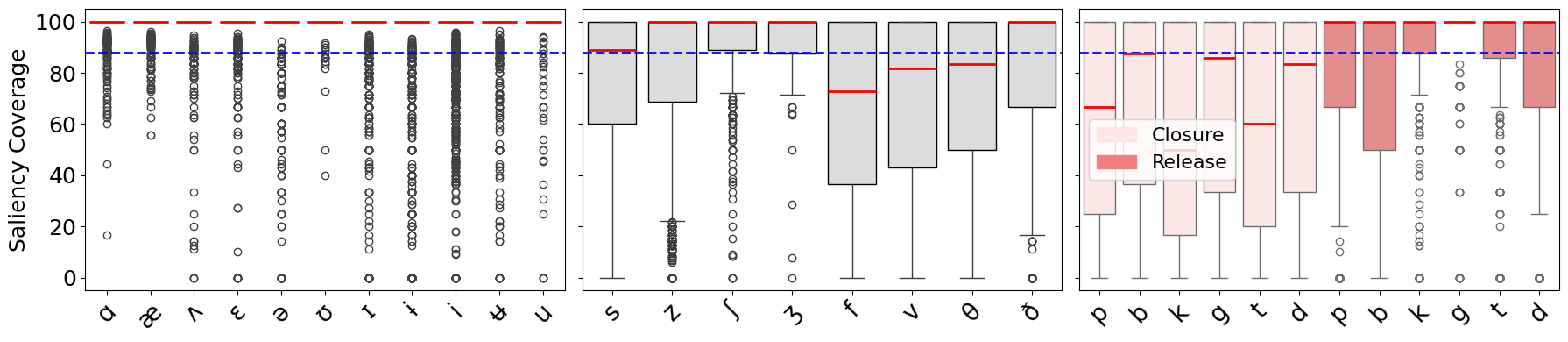
    }
    \caption{Distribution of
    time coverage scores 
    (\( \text{TC}_\phone \))
    for vowels (left), fricatives (center), and plosives (right). The dashed blue line represents the average
    coverage for a single word as a reference.}
    \label{fig:duration}
\end{figure*}

\section{Results}


\subsection{Time}
\label{sec:time_results}

Fig. \ref{fig:duration} shows the 
distributions of the \textit{time coverage} (\( \text{TC}_\phone \)) scores.
Distinct trends emerge between vowels and other phonemes. For \textbf{vowels}, the \( \text{TC}_\phone \) 
scores
are highly concentrated near $100$\% with minimal variability.
Their scores---except for outliers---exceed the average coverage for single words (used as a reference), showing that 
the model focuses more on vowel time spans than on other sounds for word prediction.
This is consistent with phonetic knowledge, which recognizes that
vowels carry rich acoustic information. Not only do they provide key cues for their identification---where the steady portion often suffices for categorization---but they also serve as contextual cues for adjacent sounds. For example, shifts between vowels and plosives or fricatives generate formant transitions that help determine the place of articulation for surrounding sounds.

Moving to \textbf{fricatives}, their \( \text{TC}_\phone \) distributions
show greater variability, with distinct patterns based on phoneme type. For sibilant fricatives (/\textipa{s}/, /\textipa{z}/, /\textipa{\textesh}/, /\textipa{\textyogh}/),
the median \( \text{TC}_\phone \) values are consistently above the average 
word-level coverage, with relatively low variability. In contrast, non-sibilant fricatives (/\textipa{f}/, /\textipa{v}/, /\textipa{\texttheta}/, /\textipa{\dh}/) 
exhibit greater variability and have medians below
the average word-level coverage
except for /\textipa{\dh}/.
Therefore, the model 
focuses more
on the time spans of sibilant fricatives than on non-sibilant fricatives. 
This can be explained by phonetic 
knowledge 
indicating 
that sibilants have well-defined spectral properties and concentrated noise energy, which likely are
more informative cues to the model than the diffuse, less acoustically distinct frication of non-sibilants \cite{maniwa_fricatives} (see also \S \ref{sec:frequency_results}).

\textbf{Plosives} 
display variable \( \text{TC}_\phone \) 
scores, 
with median values occasionally falling below the average coverage for single words. 
Comparing
closure and release phases, we observe that closure tends to have lower \( \text{TC}_\phone \) scores than release.
During the release phase, median \( \text{TC}_\phone \) values for both voiced and voiceless plosives consistently reach \( 100\% \), though their distributions vary in sparsity.
These scores suggest that release provides the model with more acoustic information than closure. 
Again, this
is consistent with phonetic knowledge, which indicates that
the release phase includes the burst,
capturing spectral properties unique to each plosive, while the closure phase is often silent and less informative for speech discrimination.



\begin{table}[t!]
\footnotesize
\centering
\setlength{\tabcolsep}{5pt} 
\caption{Spectral match (\( \text{SM}_\phoneme \)) 
for formants,
split by male (M) and female (F) speakers.}

\begin{tabular}{c||cc|cc|cc|cc}
\toprule
 & \multicolumn{2}{c|}{\textbf{F1}} & \multicolumn{2}{c|}{\textbf{F2}} & \multicolumn{2}{c|}{\textbf{F3}} & \multicolumn{2}{c}{\textbf{F4}} \\
\cline{2-9}
 & \textbf{M} & \textbf{F} & \textbf{M} & \textbf{F} & \textbf{M} & \textbf{F} & \textbf{M} & \textbf{F} \\
\hline
\textipa{\textscripta} & 94.5 & 85.7 & 91.8 & 88.7 & 43.7 & 44.4 & 24.1 & 22.1 \\
\textipa{\ae} & 96.5 & 87.9 & 97.1 & 90.7 & 60.4 & 67.7 & 27.2 & 25.8 \\
\textipa{\textturnv} & 86.7 & 75.5 & 90.2 & 88.9 & 35.9 & 36.8 & 31.3 & 35.6 \\
\textbf\textipa{\textepsilon} & 90.4 & 79.0 & 93.3 & 87.9 & 67.2 & 65.7 & 28.9 & 30.4 \\
\textipa{\textschwa} & 70.2 & 60.4 & 76.7 & 70.3 & 31.1 & 33.4 & 24.5 & 27.8 \\
\textipa{\textupsilon} & 74.5 & 73.6 & 74.5 & 64.4 & 35.1 & 34.5 & 29.3 & 27.6 \\
\textipa{\textsci} & 64.9 & 56.4 & 78.8 & 74.5 & 59.4 & 59.7 & 35.8 & 38.7 \\
\textipa{\textbari} & 54.4 & 52.0 & 69.6 & 62.3 & 42.5 & 41.3 & 25.5 & 23.7 \\
\textipa{i} & 42.7 & 36.3 & 61.1 & 54.2 & 51.8 & 58.3 & 43.0 & 37.1 \\
\textipa{\textbaru} & 55.5 & 51.7 & 52.7 & 56.9 & 50.7 & 48.0 & 29.6 & 33.5 \\
\textipa{u} & 55.2 & 54.4 & 55.2 & 57.4 & 33.3 & 29.4 & 18.4 & 17.7 \\
\hline
avg. & 71.4 & 64.8 & 73.7 & 72.4 & 46.5 & 47.2 & 28.9 & 29.1 \\
\bottomrule
\end{tabular}
\label{tab:formant}
\end{table}

\begin{table}[t!]
\footnotesize
\centering
\setlength{\tabcolsep}{5pt} 
\caption{Spectral match (\( \text{SM}_\phoneme \)) 
for peaks in fricatives/plosive releases, split by male (M) and female (F) speakers.}

\begin{tabular}{c||cc?c||cc}
\toprule
 & \textbf{M} & \textbf{F} & & \textbf{M} & \textbf{F} \\
\hline
\textipa{s} & 48.5 & 54.5 & \textipa{p} & 22.1 & 22.0 \\
\textipa{z} & 52.3 & 57.0 & \textipa{b} & 47.6 & 48.5 \\
\textipa{\textesh} & 76.4 & 75.1 & \textipa{k} & 70.0 & 72.8 \\
\textipa{\textyogh} & 63.5 & 66.7 & \textipa{g} & 69.6 & 77.6 \\
\textipa{f} & 20.3 & 8.4 & \textipa{t} & 72.5 & 78.7 \\
\textipa{v} & 12.9 & 9.1 & \textipa{d} & 60.6 & 66.2 \\
\textipa{\texttheta} & 23.4 & 17.6 & & &  \\
\textipa{\dh} & 28.9 & 28.8 & & &  \\
\hline
avg. & 40.8 & 39.7 & avg. & 57.1 & 61.0 \\
\bottomrule
\end{tabular}
\label{tab:peak}
\end{table}


\subsection{Frequency}
\label{sec:frequency_results}

\begin{figure}[t!]
    \centering
    \includegraphics[width=0.42\textwidth]{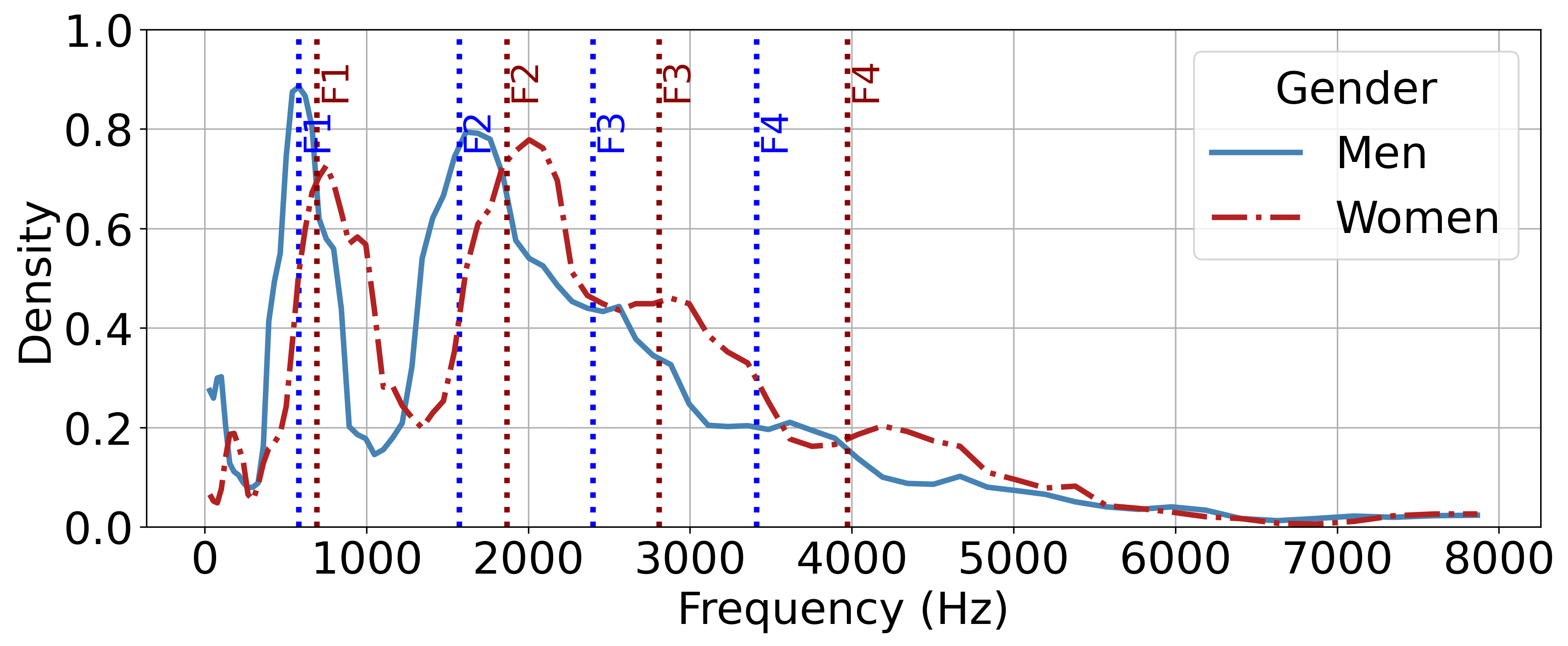}
    \caption{
    Distribution of the top 3\% most salient elements across frequencies (\( \text{D}_\phoneme \)) at the midpoint of /\textipa{\textepsilon}/, with averaged formants (F1-F4, dotted lines), by gender.}
    \label{fig:vowel_distribution}
\end{figure}

The results for the 
\textit{spectral match}
(\( \text{SM}_\phoneme \)) between salient scores and formants/spectral peaks are reported in Table~\ref{tab:formant} for vowels and Table~\ref{tab:peak} for plosives and fricatives, split by gender.
For \textbf{vowels}, the first key observation is that average scores are higher for F1 and F2 (avg. F1: $71.4$, $64.8$; avg. F2: $73.7$, $72.4$ for F and M, respectively) compared to F3 and F4 (avg. \df{F3}: $46.5$, $47.2$; avg. \df{F4}: $28.9$, $29.1$ for F and M, respectively) across both genders. 
This trend is also evident when inspecting \( \text{D}_\phoneme \), the distributions of salient elements across frequencies (see Fig. \ref{fig:vowel_distribution}),
which shows peaks corresponding to formant regions for /\textipa{\textepsilon}/.
%
When examining differences between vowels, we observe that \( \text{SM}_\phoneme \) scores for F1 and F2 increase as the openness of vowels increases:
open (/\textipa{\textscripta}/), near-open (/\textipa{\ae}/), and open-mid (/\textipa{\textturnv}/, /\textipa{\textepsilon}/) vowels show \( \text{SM}_\phoneme \) scores above $79.0$ for F1 and $87.9$ for F2. In contrast, 
mid (/\textipa{\textschwa}/) and near-close (/\textipa{\textsci}/, /\textipa{\textupsilon}/) vowels have \( \text{SM}_\phoneme \) scores below $74.5$ for F1 and $78.8$ for F2, while close vowels (/\textipa{i}/, /\textipa{\textbari}/, /\textipa{\textbaru}/, /\textipa{u}/) below $55.5$ for F1 and $61.1$ for F2. Although the exact reason for this pattern is unclear, vowel closeness leads to lower F1 values, which may cause saliency peaks to either shift toward F0 and misalign with actual F1 values (\df{see /\textipa{\textupsilon}/, /\textipa{u}/, /\textipa{\textbaru}/ in the supplementary materials) 
or simply have lower density 
(see /\textipa{\textschwa}/, /\textipa{\textsci}/, /\textipa{i}/, /\textipa{\textbari}/ in the supplementary materials)}. 
%
Additionally, \( \text{SM}_\phoneme \) scores for F1 and F2 are consistently lower for women---except for F2 in /\textipa{\textbaru}/ and /\textipa{u}/---\df{as also shown}
in the \( \text{D}_\phoneme \) distributions
(see Fig. \ref{fig:vowel_distribution} 
\df{for /\textipa{\textepsilon}/}
and 
\df{plots in the} 
supplementary materials 
\df{for other vowels}).
This suggests 
that the model more effectively captures
acoustic patterns in vowels produced by men.
\df{Future research could investigate whether gender-related differences in saliency maps align with observed gender performance gaps in ASR (e.g., \cite{attanasio-etal-2024-twists}).}
Overall, despite variations in factors like vowel openness and female speech, these findings show that a modern ASR model prioritizes F1 and F2, 
aligning to knowledge about human speech perception \cite{hillenbrand_vowels}.

\begin{figure}[t!]
    \centering
    \includegraphics[width=0.38\textwidth]{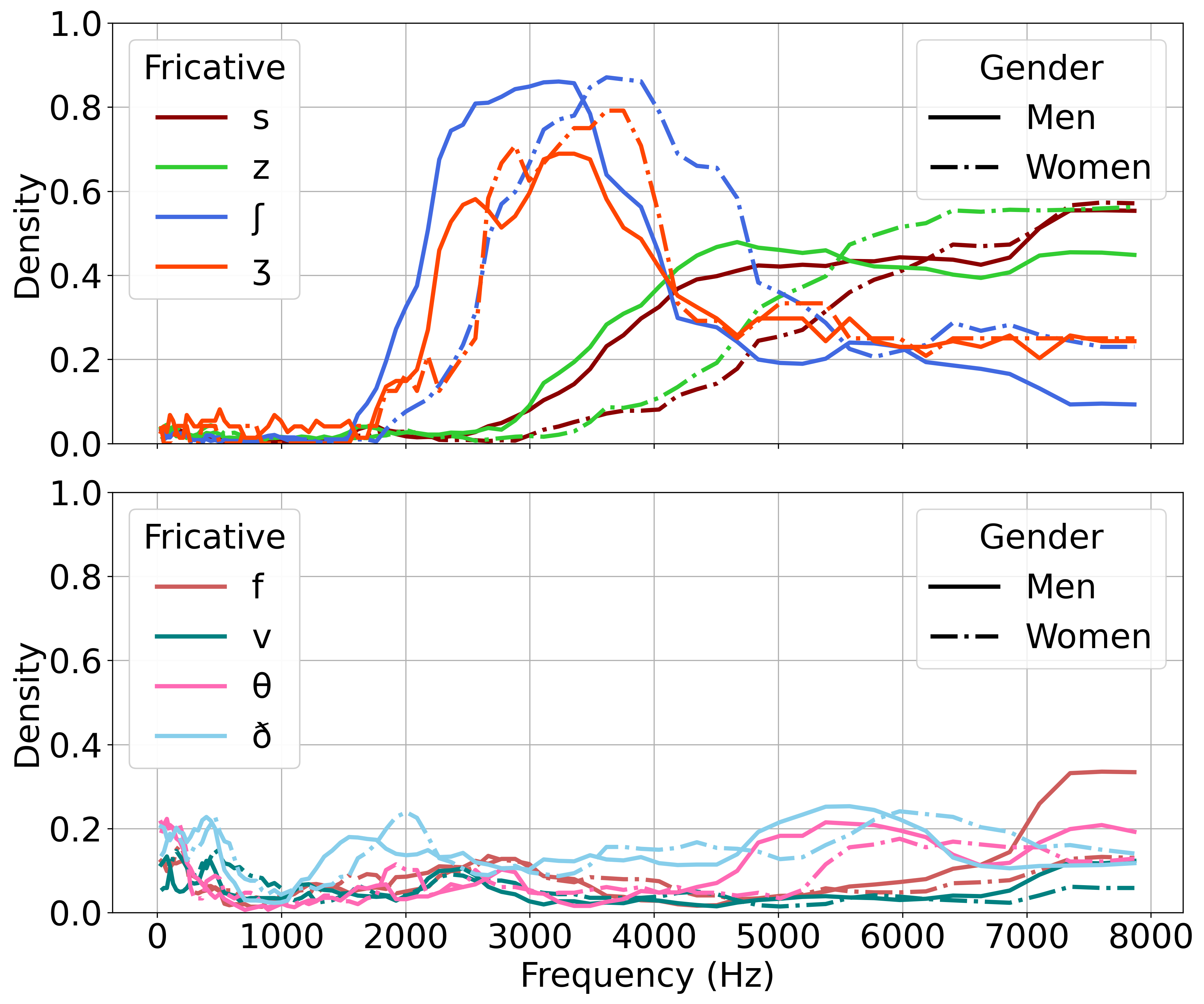}
    \caption{Distribution of the top 3\% most salient elements across frequencies (\( \text{D}_\phoneme \)) at fricative midpoint, by gender.}
    \label{fig:fricative_distribution}
\end{figure}

For \textbf{fricatives}, 
the \( \text{SM}_\phoneme \) scores vary widely, 
ranging from $8.4$ for /\textipa{f}/ to $75.1$ for /\textipa{\textesh}/ in women and from $12.9$ for /\textipa{v}/ to $76.4$ for /\textipa{\textesh}/ in men (see Table \ref{tab:peak}). On average, no substantial differences are observed between genders.
A notable trend emerges between sibilant (/\textipa{s}/, /\textipa{z}/, /\textipa{\textesh}/, /\textipa{\textyogh}/) and non-sibilant (/\textipa{f}/, /\textipa{v}/, /\textipa{\texttheta}/, /\textipa{\dh}/) fricatives, similar to the 
time
coverage results (see \S \ref{sec:time_results}): sibilants consistently yield higher scores across both genders. 
This trend is also evident in Fig. \ref{fig:fricative_distribution}, where 
salient elements are frequently located between \numprint{2000} and \numprint{4000} Hz for postalveolar fricatives (/\textipa{\textesh}/, /\textipa{\textyogh}/) and above \numprint{4000} Hz for alveolar fricatives (/\textipa{s}/, /\textipa{z}/).
These peaks are shifted toward higher frequencies for women, as expected, with minimal differences between voiced (/\textipa{z}/, /\textipa{\textyogh}/) and voiceless (/\textipa{s}/, /\textipa{\textesh}/) variants, and correspond to regions where spectral energy typically concentrates for these phonemes.
In contrast, non-sibilant fricatives lack clear 
peaks
in their saliency distributions, which are often below $0.2$ in density.
This mirrors the flatter spectra of labiodental (/\textipa{f}/, /\textipa{v}/) and dental (/\textipa{\texttheta}/, /\textipa{\dh}/) fricatives, which have no dominant frequency range, 
leading to lower \( \text{SM}_\phoneme \) scores and less defined saliency distributions.
Interestingly, this finding suggests that the model primarily relies on well-defined acoustic cues, and when these are absent, as with non-sibilants, it may activate compensatory mechanisms that warrant further investigation.

\begin{figure}[t!]
    \centering
    \includegraphics[width=0.38\textwidth]{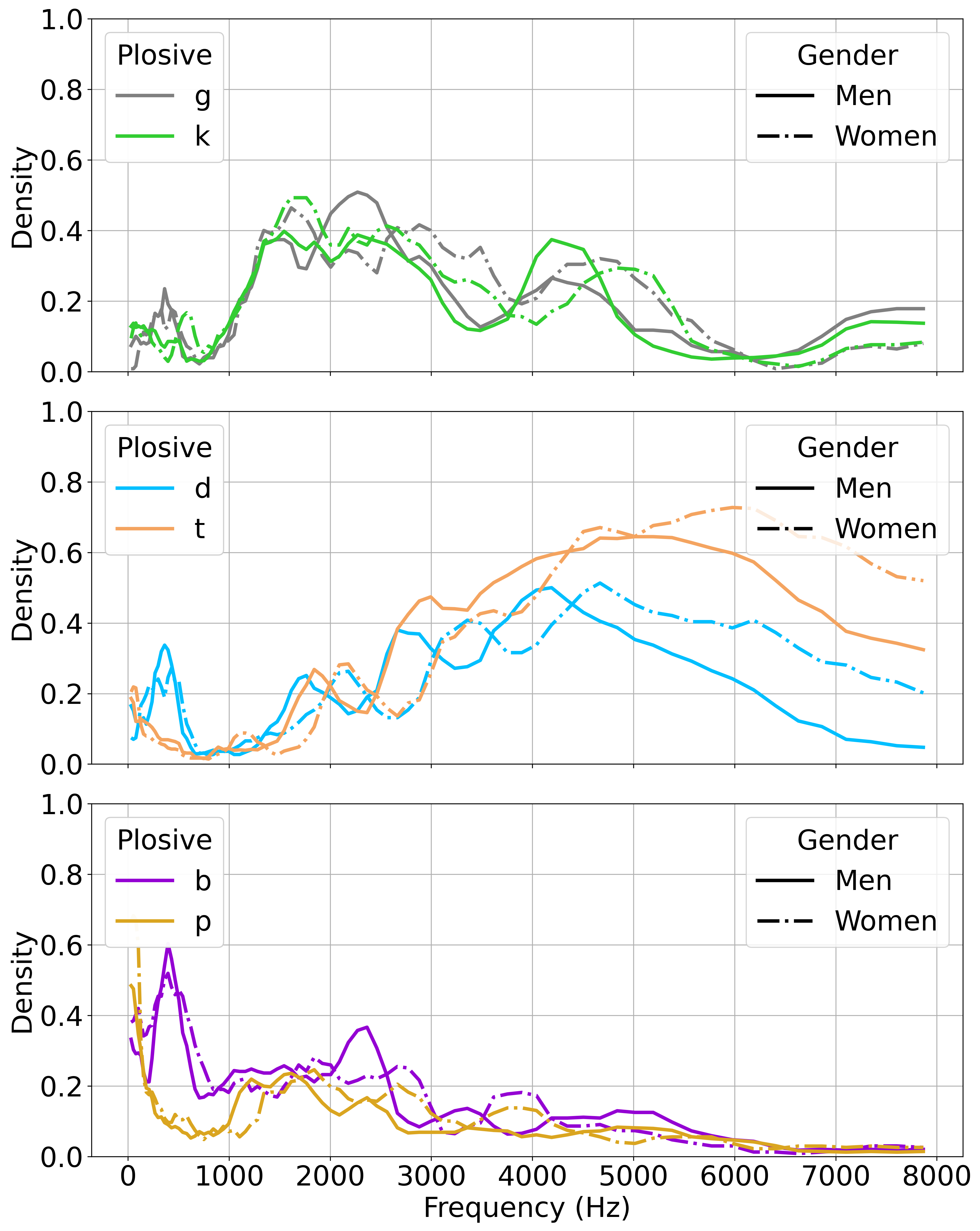}
    \caption{Distribution of the top 3\% most salient elements across frequencies (\( \text{D}_\phoneme \)) at plosive release onset, by gender.}
    \label{fig:plosive_distribution}
\end{figure}

The \( \text{SM}_\phoneme \) scores for \textbf{plosive} releases are generally between $60.6$ and $78.7$, with the exception of labials (/\textipa{b}/, /\textipa{p}/), 
whose scores range from $22.0$ to $48.5$ (see Table \ref{tab:peak}).
\( \text{D}_\phoneme \) distributions
exhibit consistent patterns across voiced (/\textipa{b}/, /\textipa{g}, /\textipa{d}/) and voiceless (/\textipa{p}/, /\textipa{k}/, /\textipa{t}/) variants (see Fig. \ref{fig:plosive_distribution}), suggesting that the model focuses on shared spectral regions for plosives with the same place of articulation, without significant gender-based differences.
Some trends align with the spectral characteristics of bursts described in the acoustic phonetics literature \cite{jackson_plosives, chodroff_burst}. 
Velar plosives (/\textipa{k}/, /\textipa{g}/) show density peaks around 0.4 between \numprint{1000} and \numprint{3000} Hz, a variation that may reflect shifts in place of articulation depending on the adjacent vowel; 
alveolar plosives (/\textipa{t}/, /\textipa{d}/) exhibit a broad density peak---from $0.4$ to $0.7$---in the high-frequency range; labial plosives (/\textipa{p}/, /\textipa{b}/) concentrate density---around $0.6$---below \numprint{1000} Hz.
In general, however, given the transient nature of burst patterns, 
\( \text{D}_\phoneme \) distributions
show 
low
overall densities, with peaks exceeding $0.5$ in density only for /\textipa{t}/, /\textipa{b}/, and /\textipa{p}/ (women).
This suggests that in the absence of well-defined acoustic cues for bursts, as in non-sibilant fricatives, the model may rely on 
more diffuse regions and, likely, complementary cues.

\section{Conclusions}

Using a feature attribution methodology, we analyzed the impact of distinctive acoustic cues of vowels, fricatives, and plosives in an English
ASR model with a Conformer-based architecture. 
\df{This is the first in-depth analysis of saliency maps in relation to fine-grained acoustic patterns across three phoneme classes.}
%
We found that, in general, the model leverages spectral cues characteristic of the phonemes analyzed, 
aligning with human speech perception.
However, the alignment between saliency scores and acoustic cues decreases as the spectral cues become less defined\df{.} 
%
Specifically: \textit{i)} the model relies heavily on vowels, using their entire span and focusing more on their F1 and F2 than F3 and F4;
\textit{ii)} the model 
focuses on more defined frequency areas in sibilant fricatives than non-sibilant fricatives;
\textit{iii)} in plosives, release is more prominent than closure, with saliency scores reflecting burst characteristics.
\df{The variability in the alignment between saliency scores and acoustic cues across sounds and genders opens avenues for future research on ASR interpretability, potentially illuminating factors affecting model robustness and performance under varying conditions.}

\noindent \textbf{Limitations}. 
Our analysis focuses on a subset of acoustic cues associated with the phonemes examined. For example, in analyzing plosives, we did not consider locus frequencies, and for fricatives, spectral moments, as these are statistical descriptors that do not have direct counterparts in the saliency maps.
Additionally, 
the generalizability of our
findings to other models and languages requires verification.
However, since most state-of-the-art ASR models use similar architectures 
and variations in the realization of the same phoneme across different languages tend to be subtle, we do not expect significant differences.

\section{Acknowledgements}

\df{The work presented in this paper has been funded from the PNRR project FAIR - Future AI Research (PE00000013),  under the NRRP MUR program funded by the NextGenerationEU, and from the European Union’s Horizon research and innovation programme under grant agreement No 101135798, project Meetween (My Personal AI Mediator for Virtual MEETings BetWEEN People).}

\bibliographystyle{IEEEtran}
\bibliography{main}

\end{document}